# An Agentic AI Framework Overcomes Fundamental Limitations of Large Language Models for Glaucoma Detection from Fundus Photography

Jalil Jalili[1,2,#], Hossein Taghizad[3,#], Anuwat Jiravarnsirikul[4],Christopher Bowd[1,2], Akram Belghith[1,2], Raheleh Kafieh[5], Christopher A. Girkin[1,2], Sally L. Baxter[1,2], Robert N. Weinreb[1,2], Linda M. Zangwill[1,2], and Mark Christopher*[1,2]

# Authors have same contribution

*Corresponding author: Mark Christopher, email: mac157@health.ucsd.edu

Author Affiliations

1. Division of Ophthalmology Informatics and Data Science, Viterbi Family Department of Ophthalmology, Shiley Eye Institute, University of California, San Diego, La Jolla, CA, USA.
2. Hamilton Glaucoma Center, Viterbi Family Department of Ophthalmology, Shiley Eye Institute, University of California, San Diego, La Jolla, CA, USA
3. Department of Electrical and Computer Engineering, University of Quebec at Trois-Rivières, Trois-Rivières, QC, Canada
4. Faculty of Medicine Siriraj Hospital, Department of Ophthalmology, Mahidol University, Bangkok, Thailand
5. Department of Engineering, Durham University, South Road, Durham DH1 3LE, UK

Funding Sources & Disclosures

**JJ**: None; **HT:** None; **RK:** None; **AB**: None; **CB**: None
**CAG:** F: National Eye Institute, EyeSight Foundation of Alabama, Research to Prevent Blindness, Heidelberg Engineering, GmbH
**SLB:** C: Topcon, F: Topcon, Optomed
**RNW:** C: Amydis, EyeGo, Iantrek, MachineMD, Remoni Health, Retispec, Spiinogenix, Topcon; F: Centervue, Optina, Machine MD, iCare, NEI, NIMHD, RPB; P: Toromedes, Zeiss Meditec
**LMZ:** F: The Glaucoma Foundation, The National Institutes of Health, The National Eye Institute; Heidelberg Engineering GmbH, Topcon Medical Systems Inc. and nonfinancial support (equipment) from Heidelberg Engineering, Carl Zeiss Meditec, Optovue, ICare Inc. Optomed Inc. and Topcon. R:EssilorLuxotica P: Zeiss Meditec AISight Health (cofounder and board member)
**MC:** O: AISight Health; F: NEI, The Glaucoma Foundation. Dr. Christopher is co-founder, inventor, board member and equity holder of AISight Health Inc.

Financial Support

NEI: R00EY030942, R01EY027510, R01EY034146 R01EY11008, P30EY022589, R01EY026590, EY022039, EY021818, R01EY023704, R01EY029058, EY19869, R21 EY027945, T35 EY033704, NIH: OT2OD032644. The Glaucoma Foundation. Unrestricted grant from Research to Prevent Blindness (New York, NY).

## Abstract

**Purpose:** Large language models (LLMs) demonstrate promising capabilities in medical image interpretation but suffer from inherent limitations including hallucination, limited diagnostic accuracy, and run-to-run inconsistency. We developed and validated an agentic AI framework that integrates LLMs with specialized deep learning tools to overcome these limitations in glaucoma detection from fundus photography.

**Methods:** We implemented an agentic AI workflow comprising three steps: (**1**) LLM initial assessment, (**2**) function calling to invoke specialized deep learning tools including quality assessment (QAModel, FundaQ-8), glaucoma classification (SwinV2-Tiny), and optic disc/cup segmentation (SegFormer-B0), and (**3**) LLM reflection that integrates the initial impression with tool outputs. We evaluated two LLMs (Gemini 2.5 Flash and GPT-5.4 mini) across two public datasets (ORIGA, n=100; RIM-ONE-v3, n=100) with two field-of-view conditions (uncropped and cropped). All images were independently reviewed by a masked fellowship-trained glaucoma specialist. Performance was assessed for glaucoma classification, cup-to-disc ratio (CDR) estimation, image quality assessment, and run-to-run consistency against both dataset labels and specialist grading.

**Results:** The agentic workflow improved glaucoma classification accuracy by 16–47 percentage points across all conditions, achieving performance within 6 percentage points of a fellowship-trained glaucoma specialist. On RIM-ONE-v3, the best agentic configurations matched the specialist's accuracy of 88%. LLM-alone approaches exhibited two distinct failure modes: GPT-5.4 mini demonstrated systematic positive bias (sensitivity 95–100%, specificity 0–5%), while Gemini 2.5 Flash showed stochastic variability between runs. The agentic workflow corrected both. CDR estimation error decreased by 15–50% (mean absolute error [MAE] from 0.156–0.228 to 0.104–0.132), and correlation with specialist grading improved from weak (r=0.12–0.39) to moderate-strong (r=0.59–0.84). Run-to-run consistency for glaucoma classification improved from near-random (κ as low as −0.01) to near-perfect agreement (κ up to 0.96).

**Conclusions:** Agentic AI frameworks that integrate LLMs with specialized deep learning tools successfully addressed some limitations of LLM-alone approaches including over-diagnosis or high run-to-run variability. The agentic approach improved results for both LLMs evaluated here, suggesting it could be generalizable across other LLM backbones. This approach provided more reliable, consistent, and accurate glaucoma predictions and could represent a paradigm shift from monolithic models toward orchestrated multi-agent systems in medical AI.

## Introduction

Glaucoma is a leading cause of irreversible blindness worldwide, affecting more than 80 million people and frequently remaining undetected until significant vision loss has occurred.[1, 2] Fundus photography is a widely accessible modality for population-level screening, but interpretation is subjective and shows substantial inter-grader variability, particularly in cup-to-disc ratio (CDR) estimation and assessment of neuroretinal rim thinning. Deep learning models trained on fundus photographs have achieved expert-level performance for glaucoma detection in benchmark settings; however, these systems are typically task-specific, lack explicit clinical reasoning, and do not provide interpretable intermediate outputs that clinicians can audit or trust.[3-5]

Large language models (LLMs) have emerged as a promising technology in medicine, demonstrating strong performance across a number of tasks including medical image interpretation, medical licensing examinations, and patient communication.[6-9] Multimodal extensions combining vision and language have enabled their application to medical imaging, including ophthalmic modalities such as fundus photography and optical coherence tomography.[10, 11] Despite promising results in experimental settings, deployment of LLMs for clinical image interpretation is constrained by three fundamental limitations.

First, LLMs lack specialized ophthalmic image grounding: their vision and language components are pretrained on natural images and broad text corpora, not on curated ophthalmic datasets, so they lack task-specific anatomical and morphometric expertise.[12, 13] As a result, they have limited ability to reliably interpret key retinal structures such as the optic disc, retinal nerve fiber layer, and vasculature, and cannot match the precision of dedicated models for quantitative tasks such as CDR estimation, as illustrated by the inconsistent and inaccurate CDR estimates across repeated evaluations shown in Figure 1A.

Second, LLM outputs are inherently non-deterministic, fundamentally limiting clinical reproducibility. In contrast to conventional deep learning systems that produce stable predictions for a given input, LLM-based image interpretation can vary substantially across runs, affecting both diagnostic labels and quantitative assessments.[13-15] In our previous work [16], repeated analysis of identical fundus images resulted in disagreement in approximately 30 to 35% of cases (Figure 1B), a level of variability that would be unacceptable in any clinical screening workflow. Notably, this inconsistency is not limited to a single model: performance varies substantially across LLM families, across versions from the same company, and across time when the same model is evaluated on identical images. [17, 18]

Third, LLMs may generate hallucinated or overconfident outputs without appropriate uncertainty calibration.[19, 20] Rather than abstaining when visual evidence is ambiguous or image quality is insufficient, they may produce plausible but unsupported anatomical descriptions or diagnostic conclusions (Figure 1C). Combined with modest classification accuracy in ophthalmic imaging studies, where fundus-based LLM performance typically ranges from 60% to 75%, these limitations raise serious concerns about their safety and reliability as standalone clinical decision support tools.[16]

Agentic AI frameworks, in which LLMs coordinate specialized external tools rather than acting as direct predictors, have recently been explored in radiology general clinical tasks.[21-23] However, systematic validation of agentic architectures for ophthalmic image analysis, and in particular their impact on run-to-run consistency as a clinical safety metric, has not been reported. To address these limitations, we designed an agentic AI framework in which the LLM functions as a clinical orchestrator rather than a standalone image interpreter. In this architecture, a structured three-step workflow comprising initial LLM assessment, function calling to invoke specialized open-source deep learning tools,[24, 25] and LLM reflection integrating tool outputs anchors predictions in deterministic, reproducible measurements while preserving the model's capacity for high-level clinical reasoning.[26, 27] Each architectural component directly targets one of the three failure modes identified above: grounding reduces hallucination, deterministic tool outputs stabilize run-to-run consistency, and reflection enables uncertainty-weighted integration of potentially conflicting signals.

In this study, we validate this framework across two multimodal LLMs (Gemini 2.5 Flash [28] and GPT-5.4 mini [29]), two public fundus datasets (ORIGA [30] and RIM-ONE-v3 [31, 32]), and two field-of-view conditions (full uncropped fundus photos and cropped optic disc-centered photos), evaluating glaucoma classification accuracy, CDR estimation, image quality assessment, and run-to-run consistency against both dataset labels and specialist grading. To our knowledge, this is the first study to demonstrate that an agentic architecture, rather than model scale or fine-tuning, can systematically correct distinct LLM failure modes and achieve performance approaching that of a fellowship-trained glaucoma specialist using cost-efficient LLM tiers.

## Methods

### *Study Design and Datasets*

This study evaluated an agentic AI framework for glaucoma detection using fundus photographs from two publicly available datasets: ORIGA and RIM-ONE-v3. For ORIGA, 100 images were randomly selected, comprised of 50 glaucoma and 50 nonglaucoma eyes based on dataset labels. For RIM-ONE-v3, all 39 available glaucoma images and 61 nonglaucoma images were included, resulting in 100 images.[16] In total, 200 fundus photographs were analyzed. Ground truth labels for both datasets were established through expert human grading: ORIGA labels followed Singapore Malay Eye Study (SiMES) clinical criteria, and RIM-ONE-v3 labels were assigned by two ophthalmologists with specialist adjudication of disagreements.

All images were independently reviewed by a fellowship-trained glaucoma specialist who was masked to dataset labels and AI predictions. The specialist graded glaucoma status (glaucoma, non-glaucoma, or cannot determine), quantitative CDR, CDR category (normal or enlarged), image gradability (gradable or ungradable), and image quality (good, moderate, or poor). Images graded as cannot determine for glaucoma status (n = 3 per dataset) were excluded from analyses comparing AI predictions with specialist assessments.For each dataset, two image versions were evaluated using the AI approaches: (1) the original fundus photograph and (2) an optic disc-centered crop. Optic disc-centered cropping was included because it is commonly applied in clinical glaucoma imaging workflows to standardize input and reduce background variation,[33] and its effect on LLM and agentic AI performance had not been previously characterized. For cropped-image experiments, an optic disc-centered square region with a width of two times the disc diameter was extracted from each fundus photograph, consistent with preprocessing used in our previous glaucoma detection studies.[3] This yielded four image sets: ORIGA-Uncropped, ORIGA-Cropped, RIM-ONE-Uncropped, and RIM-ONE-Cropped. Representative images from both datasets are shown in Figure 2.

### *Large Language Models and Agentic AI Framework*

Two multimodal LLMs were evaluated: GPT-5.4 mini (OpenAI, San Francisco, CA, USA) [29]  and Gemini 2.5 Flash (Google, Mountain View, CA, USA) [28], accessed via their respective APIs in April-May 2026. Both models were evaluated using default inference settings without modification of generation parameters, including temperature, to reflect realistic deployment conditions in which clinical users interact with standard configurations rather than tuned settings. Identical prompts and evaluation procedures were applied to both models.

For the LLM-alone workflow, each image was analyzed using a structured prompt requesting assessment of six outputs: image quality, image gradability, CDR category, quantitative CDR estimate, glaucoma status, and glaucoma probability. Structured outputs were requested to facilitate automated extraction and analysis. The full prompt templates used for the LLM-alone workflow and for the agentic AI reflection step are provided in Supplemental Figure S1.

The agentic workflow consisted of three sequential components (Figure 3), in which the LLM functions as a clinical orchestrator rather than a standalone predictor. In the first step, the LLM processed the fundus photograph using the same structured prompt as the LLM-alone workflow, producing an initial diagnostic impression. For cropped-image experiments, the cropped image was provided at this stage. In the second step, function calling enabled the LLM to invoke four specialized open-source deep learning models, applied to the full fundus photograph: QAModel for automated image quality scoring [34, 35], FundaQ-8 for clinical quality grading [36, 37], SwinV2-Tiny for glaucoma classification [38, 39], and SegFormer-B0 for optic disc and cup segmentation with vertical CDR calculation.[40, 41] Details of each deep learning tool, including training datasets, input specifications, and output formats, are provided in Table 1. The framework was implemented using native function calling, which enables the LLM to determine which tools to invoke and in what order, based on input and prompt, rather than following a fixed sequential pipeline. In the third step, the LLM received its initial impression alongside all deep learning model outputs and generated a final integrated diagnostic interpretation and structured report. All combinations of dataset (ORIGA, RIM-ONE-v3), field of view (uncropped, cropped), workflow (LLM-alone, agentic), and LLM (Gemini 2.5 Flash, GPT-5.4 mini) were evaluated, resulting in 16 experimental conditions.

### *Evaluations and Repeatability Analysis*

To evaluate run-to-run consistency, each experimental condition was performed twice using identical inputs and settings. The primary outcome was glaucoma classification accuracy relative to dataset labels, with additional metrics including sensitivity, specificity, F1-score, and Cohen's kappa (k) against both dataset labels and specialist grading. CDR estimation was evaluated across two dimensions: (**1**) categorical agreement (enlarged vs. normal) reported as accuracy, F1-score, and Cohen's k against specialist grading and (**2**) numerical agreement reported as mean absolute error (MAE) and Pearson correlation coefficient. Image quality assessment was evaluated as a three-class classification (good, moderate, or poor) against specialist grading, reported as accuracy, quadratic-weighted Cohen's k, and recall for poor-quality images, the last representing a clinically critical safety metric. Run 1 results are reported as the primary performance evaluation. Run-to-run consistency was assessed by comparing Run 1 and Run 2 outputs.

## Results

### *Glaucoma Classification Performance*

Glaucoma classification results are presented in Table 2. The agentic workflow improved classification accuracy by 16 to 47 percentage points compared to LLM-alone across all experimental conditions. On the ORIGA dataset, accuracy for Gemini 2.5 Flash improved from 57% (LLM-alone) to 73% (agentic) with uncropped images and from 55% to 73% with cropped images. For GPT-5.4 mini, accuracy improved from 62% to 72% (uncropped) and from 50% to 69% (cropped). On the RIM-ONE-v3 dataset, the agentic approach demonstrated even larger improvements. For Gemini 2.5 Flash, accuracy improved from 61% to 85% (uncropped) and from 54% to 88% (cropped). For GPT-5.4 mini, accuracy improved from 38% to 85% (uncropped) and from 40% to 87% (cropped).

Sensitivity and specificity varied substantially across LLMs and datasets under the LLM-alone workflow. GPT-5.4 mini achieved high sensitivity (95–100%) but near-zero specificity (0–5%) on three of four conditions (ORIGA cropped, RIM-ONE uncropped, RIM-ONE cropped), while Gemini 2.5 Flash demonstrated more variable patterns with sensitivity ranging from 40% to 87% and specificity from 36% to 74%. With the agentic workflow, less variation in sensitivity was observed for both LLMs (sensitivity range:74–90%, specificity range:40–93%). For Gemini 2.5 Flash on uncropped ORIGA images, sensitivity increased from 40% (LLM-alone) to 76% (agentic), while specificity decreased modestly from 74% to 70%. For GPT-5.4 mini on uncropped RIM-ONE, sensitivity decreased from 97% to 77% while specificity increased from 0% to 90%. F1 scores improved from 0.48–0.69 (LLM-alone) to 0.73–0.83 (agentic) across all conditions. A complete breakdown of accuracy, sensitivity, specificity, and F1-score for all comparisons is provided in Supplemental Figure S2.

The agentic AI performance approached that of the fellowship-trained glaucoma specialist on both datasets. On ORIGA, the specialist achieved 79% accuracy (sensitivity 76%, specificity 83%, F1 0.79), while the best agentic configuration achieved 73% accuracy. On RIM-ONE-v3, the specialist achieved 88% accuracy (sensitivity 85%, specificity 90%, F1 0.85), closely matched by the agentic configurations achieving 85–88% accuracy. Field of view affected the two workflows differently: LLM-alone performance varied between cropped and uncropped images, whereas agentic performance was stable across both conditions.

### *Agentic AI vs. DL tools*

The agentic workflow not only corrected LLM errors but also overrode incorrect deep learning predictions when evidence conflicted. On ORIGA, the DL classifier (SwinV2-Tiny) achieved 68% accuracy, with high sensitivity (0.98) but low specificity (0.38) from 31 false positives, while the agentic workflow achieved 69–73%, correcting 2–17 classifier false positives. On RIM-ONE-v3, where the DL classifier achieved 88% accuracy (sensitivity 0.79, specificity 0.93), the agentic workflow achieved similar performance (85–88%). Figure 4 illustrates two ways the workflow corrected errors in these datasets, although the models were not explicitly trained or constrained to do so: correction of LLM errors using DL tool outputs (Figure 4A) and resolution of conflicting DL outputs (Figure 4B). Analysis of the deep learning pipeline revealed that the segmentation-derived CDR category (enlarged vs. normal) the glaucoma classifier (SwinV2-Tiny) agreed in 80.6% of cases, where the agentic AI followed their consensus 96.1% of the time. When the segmentation and classification models disagreed (19.4% of cases), the agent arbitrated between them. Its decisions were more accurate when it deferred to the segmentation-derived CDR (71.7%) than to the classifier (54.5%). Representative outputs from all four deep learning tools, including quality scores, glaucoma probability, and optic disc/cup segmentation with estimated CDR, are shown for four illustrative cases in Supplemental Figure S3.

### *Agreement with Specialist and Run-to-Run Consistency*

Agreement between AI predictions and specialist grading and run-to-run consistency is presented in Table 3. For agreement with the specialist, LLM-alone approaches achieved Cohen's κ values ranging from -0.03 to 0.36, representing no to fair agreement per Landis–Koch criteria.[42] The agentic workflow substantially improved specialist agreement, with κ values ranging from 0.38 to 0.84, representing fair to almost perfect agreement. On ORIGA, Gemini improved from κ = 0.07–0.36 (LLM-alone) to κ = 0.38–0.55 (agentic), and GPT-5.4 mini improved from κ = 0.00–0.21 (LLM-alone) to κ = 0.39–0.48 (agentic). On RIM-ONE, the improvements were larger: Gemini improved from κ = 0.11–0.25 (LLM-alone) to κ = 0.80–0.84 (agentic), and GPT-5.4 mini improved from κ = -0.03–0.01 to κ = 0.77–0.83.

Run-to-run consistency, measured as agreement between repeated analyses, showed marked improvement with the agentic workflow. LLM-alone approaches achieved run-to-run κ values ranging from -0.01 to 0.51, with 25–50% of classifications changed between identical runs. With the agentic workflow, run-to-run κ values improved to 0.40–0.96. The most dramatic improvement was observed for GPT-5.4 mini on uncropped RIM-ONE-v3, where run-to-run κ improved from -0.01 (LLM-alone) to 0.85 (agentic). The highest consistency was achieved by GPT-5.4 mini on RIM-ONE-v3-Cropped with the agentic workflow (κ = 0.96).

### *Cup-to-Disc Ratio Estimation*

Results for CDR classification and quantitative estimation are presented in Table 4. For categorical CDR classification (normal vs. enlarged), the agentic workflow improved accuracy from 46–65% (LLM-alone) to 73–90% across all conditions, with F1 improving from 0.62–0.76 to 0.73–0.88. For quantitative estimation, mean absolute error (MAE) decreased from 0.156–0.228 to 0.104–0.132, and Pearson correlation with specialist grading improved from r = 0.12–0.39 to r = 0.59–0.84. LLM-alone correlation was higher for uncropped than cropped images (r = 0.28–0.39 vs. 0.12–0.28), a preprocessing dependency the agentic workflow eliminated (r = 0.62–0.84 vs. 0.59–0.83), as illustrated in Supplemental Figure S4.

Run-to-run consistency for CDR estimation is summarized in Supplemental Table S1. Categorical CDR consistency improved from κ = 0.00–0.66 (LLM-alone) to κ

= 0.51–0.90 (agentic), and numerical between-run MAE for Gemini decreased from 0.103–0.141 to 0.019–0.053. As in Table 2, GPT-5.4 mini reached κ = 0 on two conditions by classifying all images as enlarged in both runs, yielding 100% observed agreement but no classification variability.

*Image Quality Assessment*

Three-class image quality assessment (good/moderate/poor) performance is presented in Table 5. The specialist graded 35 images as good, 52 as moderate, and 13 as poor in ORIGA (n=100), and 11 as good, 71 as moderate, and 18 as poor in RIM-ONE-v3 (n=100). Overall classification accuracy ranged from 29% to 70%, with the highest accuracy achieved by GPT-5.4 mini LLM-alone on uncropped RIM-ONE-v3 images (70%). Quadratic-weighted κ values ranged from 0.08 to 0.50, indicating slight to moderate agreement with specialist grading.

Detection of poor-quality images, a safety-critical capability for clinical deployment, showed marked variation across conditions. On ORIGA with uncropped images, Gemini LLM-alone failed to detect any poor-quality images (0/13, 0%), while GPT-5.4 mini LLM-alone detected only one (1/13, 8%). The agentic workflow improved poor-quality detection to 31% (4/13) for Gemini and 23% (3/13) for GPT-5.4 mini on this condition. On RIM-ONE-v3, the agentic workflow achieved 100% poor-quality recall across all conditions for both LLMs, compared to 56–94% for LLM-alone approaches. However, improved poor-quality detection on RIM-ONE-v3 was accompanied by decreased overall accuracy and quadratic κ. For Gemini with uncropped images, accuracy decreased from 50% (LLM-alone) to 39% (agentic), and quadratic κ decreased from 0.42 to 0.10, suggesting that the agentic workflow may over-classify images as poor quality on this dataset.

Run-to-run consistency for quality assessment showed variable patterns. On ORIGA with uncropped images, run-to-run κ improved from 0.53 (Gemini LLM-alone) to 0.85 (agentic). On RIM-ONE-v3, run-to-run consistency was generally lower for the agentic workflow compared to LLM-alone in several conditions.

## Discussion

This study demonstrates that agentic AI architectures can address limitations of LLM-alone approaches for ophthalmic image analysis. The agentic workflow improved glaucoma classification accuracy, reduced CDR estimation error, and improved run-to-run consistency. Results were consistent across underlying LLMs, datasets, and fundus field-of-view conditions. These improvements suggest agentic approaches can address three limitations that currently prevent reliable clinical deployment of LLMs for medical imaging: hallucination, limited accuracy, and inconsistency.

The performance gains can be attributed to three architectural features. First, using deep learning models specialized for specific tasks (glaucoma classification, disc/cup segmentation, image quality assessment) to help ground model predictions can help reduce hallucination. These deep learning models provide external signals that anchor LLM interpretation to their validated predictions. Second, these deterministic tools also help increase consistency. LLM outputs are stochastic: text generation introduces randomness at each step, so identical inputs can yield different outputs across runs. The deep learning models in this pipeline, by contrast, provide deterministic predictions. Identical inputs produce identical outputs, reducing run-to-run variability. Third, the reflection step integrates potentially conflicting signals through uncertainty weighting, disagreement resolution, and quality-aware interpretation. The final assessment synthesizes evidence from both the LLM's initial impression and the specialized tool outputs.

Prior efforts to apply LLMs directly to fundus images for glaucoma detection remain limited and report consistently modest, highly variable performance. Using GPT-4V on ORIGA and RIM-ONE-v3, we previously observed accuracies of 0.70 and 0.81, respectively, below expert graders and with substantial run-to-run inconsistency.[16] On the REFUGE dataset, ChatGPT-4 reached 90% accuracy but only 50% sensitivity, reflecting a strong bias toward classifying images as normal,[43] and on ACRIMA, prompt-engineered evaluations of GPT-4o and Claude Sonnet 3.5 yielded F1 scores of only 0.65 and 0.73.[44] When restricted to structured-text prompts rather than direct image input, GPT-4 performed poorly on glaucoma referral specifically (F1 0.04).[45] Across these studies, accuracy varies not only with model family and version but with prompt design, confirming that raw LLM performance on this task is inconsistent.

The agentic workflow improved glaucoma classification accuracy across all experimental conditions. The largest gains were observed for GPT-5.4 mini on RIM-ONE-v3 (38% to 85% uncropped; 40% to 87% cropped). Both LLMs were evaluated with identical prompts and default inference settings, isolating the agentic architecture as the source of performance gains. On RIM-ONE, the best agentic configurations matched the specialist's accuracy of 88%. This near-specialist performance was achieved using cost-efficient LLM tiers (Gemini 2.5 Flash and GPT-5.4 mini) rather than flagship models. This suggests that architectural design, rather than raw model scale, is the primary driver of the performance gains observed here.

The reflection step is not merely a pass-through mechanism. On ORIGA, the agentic workflow achieved higher accuracy than the DL classifier alone, correcting classifier false positives (illustrated in Figure 4A, where reflection reversed the initial LLM classification from non-glaucoma to glaucoma) by recognizing that normal CDR estimates from the segmentation model were inconsistent with the classifier's high glaucoma probability. The reflection step can also resolve discrepancies between deep learning tools themselves. When the classifier and segmentation model produce conflicting outputs, the LLM applies clinical reasoning to arbitrate, prioritizing morphological evidence over isolated probability scores. The agentic workflow therefore functions as an intelligent integrator rather than a simple aggregator of tool outputs. It does not eliminate variability entirely; the LLM reflection step still introduces some stochastic variation.

LLM-alone approaches exhibited systematic classification biases. GPT-5.4 mini demonstrated near-complete positive bias, classifying nearly all images as glaucoma regardless of true disease status. This pattern would result in unacceptably high false-positive rates in a screening setting, leading to unnecessary referrals and patient anxiety. Gemini 2.5 Flash, in contrast, showed low sensitivity in certain conditions, missing the majority of glaucoma cases, a critical failure for a screening tool intended to identify disease. The agentic workflow corrected both biases by grounding LLM predictions in the outputs of specialized deep learning models, achieving balanced sensitivity and specificity that more closely approximated specialist performance. This transition from biased to balanced classification represents a fundamental improvement in clinical utility. The two LLMs failed in distinct ways, one through systematic positive bias and the other through stochastic variability, yet the agentic architecture corrected both and achieved comparable final performance across LLM families. The benefits of agentic AI therefore appear generalizable rather than dependent on a specific LLM backbone. The agentic workflow was also robust to field of view. LLM-alone performance varied between cropped and uncropped images, whereas agentic performance was stable across both, because the deep learning tools always received the full fundus image.

The agentic AI also resolved conflicts between the DL classifier and segmentation model. When these models disagreed, for example, when the classifier predicted glaucoma while the CDR remained below 0.5, the system did not default to either source. Instead, it relied on the LLM's clinical assessment of disc morphology, image quality, and rim appearance during the reflection step. Accuracy was higher when following segmentation-derived CDR than the classifier (71.7% vs. 54.5%). This suggests that morphological features may provide more reliable guidance than probability scores when DL models conflict, and that the agentic framework can effectively leverage this signal through structured reasoning.

Run-to-run consistency is a critical requirement for clinical deployment that has received limited attention in prior evaluations of LLMs for medical imaging. A diagnostic system that produces different outputs when analyzing the same image twice cannot be reliably integrated into clinical workflows. LLM-alone approaches exhibited substantial stochastic variability, with run-to-run agreement as low as $\kappa = -0.01$ for GPT-5.4 mini on RIM-ONE-v3, effectively random variation between identical analyses. This inconsistency likely reflects the inherent non-determinism of LLM inference, where sampling-based text generation introduces variability even with identical inputs. The agentic workflow improved consistency dramatically (κ up to 0.96), an architectural consequence of grounding the final interpretation in deterministic deep learning model outputs rather than relying solely on stochastic LLM generation. The improvement in specialist agreement was particularly notable on RIM-ONE, where GPT-5.4 mini improved from near-zero agreement (κ = -0.03 to 0.01) to substantial agreement (κ = 0.77-0.83), a transition from chance-level to expert-level concordance. Both runs of the agentic workflow achieved similar agreement with the specialist (e.g., Gemini on RIM-ONE-v3: Run 1 κ = 0.81, Run 2 κ = 0.80), indicating that this improvement is reproducible rather than a chance finding from a single favorable run.

While cup-to-disc ratio has limited clinical utility on its own, it remains widely used and provides a quantitative output well suited to testing LLM numerical accuracy. LLM-alone approaches demonstrated limited precision, with MAE values of 0.156-0.228. These represent substantial errors relative to the 0–1 CDR scale. The agentic workflow reduced MAE by 15-50% and improved Pearson correlation from weak (r = 0.12-0.39) to moderate-strong (r = 0.59-0.84). This improvement is attributable to the SegFormer-based optic disc and cup segmentation model, which provides precise anatomical delineation rather than relying on the LLM's internal visual estimation. The categorical CDR κ = 0.00 observed for GPT-5.4 mini on two conditions reflects complete classification invariance: the model classified every image as "Enlarged" regardless of true CDR status. The agentic workflow corrected this systematic bias by grounding decisions in objective anatomical measurements, achieving κ values up to 0.90. LLM-alone CDR estimation showed higher correlation with specialist grading on uncropped fundus images than on cropped images. The agentic workflow eliminated this preprocessing dependency, because the deep learning segmentation model always received the full fundus image.

Image quality assessment is a critical safety function: AI systems must recognize when image quality precludes reliable interpretation. Gemini LLM-alone failed to identify any poor-quality images in ORIGA with uncropped images (0/13, 0%). The LLM generated confident diagnostic outputs for images a specialist deemed unsuitable for interpretation, a form of hallucination (Figure 1C). The agentic workflow improved poor-quality detection on ORIGA (0-8% to 23-31%). On RIM-ONE-v3, however, the deep learning quality models over-classified images as poor. This may reflect training on fundus photographs with different illumination characteristics than the darker RIM-ONE-v3 images, although QAModel's training data are not publicly documented and this hypothesis cannot be directly tested.

Beyond quality classification accuracy, a central question is whether the agentic workflow reduces hallucinated diagnostic confidence on images that should not be confidently interpreted. The LLM-alone example in Figure 1C illustrates this risk concretely: a fundus image rated ungradable by the specialist nonetheless received a confident non-glaucoma classification with 95% probability from the LLM, with no indication of uncertainty. In the agentic workflow, this type of hallucination is constrained by two mechanisms. First, the quality assessment tools (QAModel, FundaQ-8) provide an independent, deterministic signal of image gradability that the LLM can weigh during reflection, rather than relying solely on its own visual impression. Second, as illustrated in Supplemental Figure S3, even when deep learning tools produce confident outputs on poor-quality images, the reflection step has access to explicit quality flags alongside these outputs, enabling the LLM to contextualize tool confidence in light of acknowledged image limitations. The over-classification of RIM-ONE-v3 images as poor quality under the agentic workflow, while reducing overall quality classification accuracy, may in this sense reflect increased systematic caution rather than a purely undesirable tradeoff: an agentic system that more readily flags uncertainty, even at the cost of some specificity, is preferable to one that produces confident, ungrounded interpretations of inadequate images. Direct measurement of hallucination reduction, such as confidence calibration or abstention rates on ungradable images, remains an important direction for future work.

The current agentic framework integrates four deep learning tools for quality assessment, optic disc/cup segmentation, and glaucoma classification. Its principal strength lies in its modular architecture: new specialized tools can be incorporated without rebuilding the underlying system. This extensibility enables several directions for future development. The general-purpose LLM backbone could be replaced with a fine-tuned ophthalmic model to improve baseline visual interpretation. The tool suite could be expanded to extract richer optic disc features such as neuroretinal rim thinning, disc hemorrhages, and RNFL defects, findings that currently require subjective LLM interpretation. The framework could also incorporate multimodal clinical data, including optical coherence tomography, visual field testing, and electronic health record information, to more closely approximate comprehensive clinical assessment. The ultimate goal is an agentic architecture that scales with clinical workflow complexity, orchestrating an expanding array of specialized tools while maintaining the interpretability and reasoning capabilities of the LLM coordinator.

The agentic framework incurs modestly higher computational cost than an LLM-alone approach. Each image is processed through three LLM passes, initial assessment, function-calling to invoke the deep learning tools, and reflection on the returned outputs, together with inference from four lightweight deep learning models. Because these tools are compact open-source models that run locally in under a second and the LLM tiers used are inexpensive (for example, Gemini 2.5 Flash at $0.30 and $2.50 per million input and output tokens), the additional cost amounts to a fraction of a cent per image. This overhead is negligible relative to the clinical cost of a missed diagnosis or an unnecessary referral, and the local, open-source nature of the tool suite avoids recurring per-call API fees, supporting practical large-scale deployment.

This study has several limitations. Our evaluation was conducted on two public benchmark datasets (ORIGA and RIM-ONE-v3), which may not fully represent the diversity of imaging conditions, patient populations, and disease severity encountered in clinical practice; external validation across multiple clinical sites and imaging platforms is needed to establish generalizability. We evaluated only two cost-efficient LLMs (Gemini 2.5 Flash and GPT-5.4 mini); performance may differ with flagship models or other LLM architectures. We focused on cup-to-disc ratio because a validated open-source segmentation model was available and because it provides a continuous variable well suited to testing LLM numerical accuracy, not because of its clinical value. CDR has substantial limitations as a biomarker; it is highly dependent on optic disc size and should not be used alone for diagnosis or progression, so our use should be understood as a test of measurement fidelity rather than an endorsement of its diagnostic value. Comprehensive structural assessment requires additional features such as disc area, rim quantification (e.g., BMO-MRW), RNFL thickness, disc hemorrhages, and peripapillary atrophy, which the modular architecture can readily accommodate. Our ground truth relied on dataset labels and a single specialist grader; future work should incorporate multiple expert graders to assess inter-observer variability and establish consensus labels. We evaluated the system on static, single-timepoint fundus photographs, whereas clinical glaucoma management requires longitudinal assessment and integration of functional testing. Finally, this study assessed technical performance metrics rather than clinical outcomes. Prospective validation examining the impact on diagnostic accuracy, referral patterns, and patient outcomes in real-world screening settings is essential before clinical deployment.

In conclusion, this study demonstrates that agentic AI architectures substantially overcome the fundamental limitations of LLM-alone approaches for ophthalmic image analysis. By integrating large language models with specialized deep learning tools through a structured workflow of initial assessment, function calling, and reflection, the agentic framework markedly improved glaucoma classification accuracy, reduced CDR estimation error, and transformed run-to-run consistency from near-random to near-perfect agreement. The agentic approach corrected two distinct LLM failure modes, systematic positive bias and stochastic variability, achieving performance approaching that of a fellowship-trained glaucoma specialist using cost-efficient LLM tiers. These findings support a paradigm shift in medical AI development: rather than relying on ever-larger monolithic models, clinically reliable systems can be achieved through thoughtful architectural design that leverages the complementary strengths of general-purpose LLMs for reasoning and specialized deep learning models for domain-specific measurement. With prospective validation and expanded tool integration, agentic AI frameworks offer a practical and scalable path toward trustworthy AI-assisted ophthalmic diagnosis.

## References:

1. Tham, Y.-C., et al., *Global prevalence of glaucoma and projections of glaucoma burden through 2040: a systematic review and meta-analysis.* Ophthalmology, 2014. **121**(11): p. 2081-2090.
2. Flaxman, S.R., et al., *Global causes of blindness and distance vision impairment 1990–2020: a systematic review and meta-analysis.* The Lancet Global Health, 2017. **5**(12): p. e1221-e1234.
3. Christopher, M., et al., *Performance of deep learning architectures and transfer learning for detecting glaucomatous optic neuropathy in fundus photographs.* Scientific reports, 2018. **8**(1): p. 16685.
4. Orlando, J.I., et al., *Refuge challenge: A unified framework for evaluating automated methods for glaucoma assessment from fundus photographs.* Medical image analysis, 2020. **59**: p. 101570.
5. Jalili, J., et al., *Performance of General-Purpose Vision Language Models and Ophthalmology Foundation Models in Glaucoma Detection and Function Prediction.* Translational Vision Science & Technology, 2025. **14**(11): p. 31-31.
6. Singhal, K., et al., *Large language models encode clinical knowledge.* Nature, 2023. **620**(7972): p. 172-180.
7. Chen, S.F., et al., *LLM-assisted systematic review of large language models in clinical medicine.* Nature medicine, 2026: p. 1-8.
8. JALILI, J., et al., *An Agentic AI Workflow for Orchestrated Fundus Image Analysis and Explainable Glaucoma Detection.* Investigative Ophthalmology & Visual Science, 2026. **67**(7): p. 1835-1835.
9. Chen, X., et al., *FFA-GPT: an automated pipeline for fundus fluorescein angiography interpretation and question-answer.* NPJ digital medicine, 2024. **7**(1): p. 111.
10. Jalili, J., et al., *Image-Quality–Aware Multimodal AI for Automated Structured OCT Report Generation in Glaucoma Evaluation.* Ophthalmology Science, 2026: p. 101254.
11. Jalili J, H.A., Mehta NN, Ali AL, Morsy MS, Gavhane Y, Wen B, Bartsch DU, Baxter SL, Weinreb RN, Zangwill LM, Freeman W, Christopher M, *Automated Clinician-Style Retinal OCT Report Generation Using a Fine-Tuned Multimodal Large Language Model: Development and Masked Clinical Evaluation.* SSRN Preprint, 2026.
12. Li, Z., et al., *Visionunite: A vision-language foundation model for ophthalmology enhanced with clinical knowledge.* IEEE Transactions on Pattern Analysis and Machine Intelligence, 2025.
13. Carlà, M.M., et al., *Advanced analysis of leading large language models for diagnostic accuracy in retinal imaging.* British Journal of Ophthalmology, 2026.
14. Yaghy, A., et al. *Large language models in ophthalmology: potential and pitfalls*. in *Seminars in Ophthalmology*. 2024. Taylor & Francis.
15. Subedi, K., *The reliability of llms for medical diagnosis: An examination of consistency, manipulation, and contextual awareness.* arXiv preprint arXiv:2503.10647, 2025.
16. Jalili, J., et al., *Glaucoma detection and feature identification via GPT-4V fundus image analysis.* Ophthalmology Science, 2025. **5**(2): p. 100667.
17. Chen, L., M. Zaharia, and J. Zou, *How is ChatGPT's behavior changing over time?* Harvard Data Science Review, 2024. **6**(2).
18. Zhang, Z., et al., *Evaluating large language models in ophthalmology: systematic review.* Journal of Medical Internet Research, 2025. **27**: p. e76947.
19. Kim, Y., et al., *Medical hallucinations in foundation models and their impact on healthcare.* arXiv preprint arXiv:2503.05777, 2025.
20. Kedia, N., et al., *ChatGPT and beyond: an overview of the growing field of large language models and their use in ophthalmology.* Eye, 2024. **38**(7): p. 1252-1261.
21. Bluethgen, C., et al., *Agentic Systems in Radiology: Design, Applications, Evaluation, and Challenges.* arXiv preprint arXiv:2510.09404, 2025.
22. Qiu, J., et al., *LLM-based agentic systems in medicine and healthcare.* Nature Machine Intelligence, 2024. **6**(12): p. 1418-1420.
23. Khosravi, B., et al., *Agentic AI in radiology: evolution from large language models to future clinical integration.* Radiology: Artificial Intelligence, 2026. **8**(2): p. e250651.
24. Patil, S.G., et al. *The berkeley function calling leaderboard (bfcl): From tool use to agentic evaluation of large language models*. in *Forty-second International Conference on Machine Learning*. 2025.
25. Kim, S., et al., *An llm compiler for parallel function calling.* arXiv preprint arXiv:2312.04511, 2023.
26. Sibai, N., et al., *The Path Ahead for Agentic AI: Challenges and Opportunities.* arXiv preprint arXiv:2601.02749, 2026.
27. Renze, M. and E. Guven, *Self-reflection in llm agents: Effects on problem-solving performance.* arXiv preprint arXiv:2405.06682, 2024.
28. Comanici, G., et al., *Gemini 2.5: Pushing the frontier with advanced reasoning, multimodality, long context, and next generation agentic capabilities.* arXiv preprint arXiv:2507.06261, 2025.
29. *Introducing GPT-5.4 mini and nano*. March 17, 2026; Available from: https://openai.com/index/introducing-gpt-5-4-mini-and-nano/.
30. Zhang, Z., et al. *Origa-light: An online retinal fundus image database for glaucoma analysis and research*. in *2010 Annual international conference of the IEEE engineering in medicine and biology*. 2010. IEEE.

31. Fumero, F., et al. *RIM-ONE: An open retinal image database for optic nerve evaluation*. in *2011 24th international symposium on computer-based medical systems (CBMS)*. 2011. IEEE.
32. Batista, F.J.F., et al., *Rim-one dl: A unified retinal image database for assessing glaucoma using deep learning.* Image Analysis and Stereology, 2020. **39**(3): p. 161-167.
33. Fan, R., et al., *Detecting glaucoma from fundus photographs using deep learning without convolutions: transformer for improved generalization.* Ophthalmology science, 2023. **3**(1): p. 100233.
34. Szegedy, C., et al. *Rethinking the inception architecture for computer vision*. in *Proceedings of the IEEE conference on computer vision and pattern recognition*. 2016.
35. Xie, Z. *fundus_image_quality*. Available from: https://github.com/ZiqianXie/fundus_image_quality.
36. Zun, L.Q., et al. *FundaQ-8: A Clinically-Inspired Scoring Framework for Automated Fundus Image Quality Assessment*. in *2025 IEEE 7th Symposium on Computers & Informatics (ISCI)*. 2025. IEEE.
37. Zun, L.Q. *FundaQ-8*. Available from: https://github.com/qmed-asia/FundaQ-8.
38. Sun, X. *swinv2_tiny_for_glaucoma_classification*. Available from: https://huggingface.co/pamixsun/swinv2_tiny_for_glaucoma_classification.
39. Liu, Z., et al. *Swin transformer v2: Scaling up capacity and resolution*. in *Proceedings of the IEEE/CVF conference on computer vision and pattern recognition*. 2022.
40. Xie, E., et al., *SegFormer: Simple and efficient design for semantic segmentation with transformers.* Advances in neural information processing systems, 2021. **34**: p. 12077-12090.
41. Sun, X. *segformer_for_optic_disc_cup_segmentation*. Available from: https://huggingface.co/pamixsun/segformer_for_optic_disc_cup_segmentation.
42. Landis, J.R. and G.G. Koch, *The measurement of observer agreement for categorical data.* biometrics, 1977: p. 159-174.
43. AlRyalat, S.A., A.M. Musleh, and M.Y. Kahook, *Evaluating the strengths and limitations of multimodal ChatGPT-4 in detecting glaucoma using fundus images.* Frontiers in Ophthalmology, 2024. **4**: p. 1387190.
44. Agbareia, R., et al., *The role of prompt engineering for multimodal LLM glaucoma diagnosis.* medRxiv, 2024: p. 2024.10. 30.24316434.
45. Tabuse, C.L., et al., *Evaluating Large Language Models for Multimodal Simulated Ophthalmic Decision-Making in Diabetic Retinopathy and Glaucoma Screening.* arXiv preprint arXiv:2507.01278, 2025.

## Tables and Figures:

| Table 1. Deep learning models integrated within the agentic AI framework for image quality verification and structural glaucoma assessment. | | | | | | |
|---|---|---|---|---|---|---|
| **Tool** | **Architecture** | **Task** | **Training Dataset** | **Input** | **Output** | **Reference** |
| QAModel | Inception V3 with custom regression head (2048→1024→512→1, sigmoid activation) | Continuous fundus image quality regression and categorical triage | Custom-labeled clinical dataset (not publicly documented) | Color fundus photograph (RGB, 299 ´ 299 ´ 3) | Bounded quality score (0.0–1.0); categorized into Low, Average, or good thresholds | Custom model; base architecture via Szegedy et al., 2016 (arXiv:1512.00567) [34]<br><br>Source code and model: https://github.com/ZiqianXie/fundus_image_quality |
| FundaQ-8 | ResNet-18 regression framework | Multi-attribute clinical image quality scoring across 8 parameters | Multi-source dataset (n = 1,800) from clinical and Kaggle sources; validated on EyeQ | Color fundus photograph (RGB, 224 ´ 224 ´ 3) | Bounded continuous quality score (0.0–1.0) | Lee et al., 2025 IEEE [36];<br><br>Source code and model: https://github.com/qmed-asia/FundaQ-8 |
| SwinV2-Tiny | Swin Transformer V2 (Tiny variant) | Binary classification of glaucomatous vs. normal optic discs | REFUGE Challenge dataset (n = 1,200) | Color fundus photograph (RGB) | Softmax probability distribution for [normal, glaucoma] classes | base architecture via Liu et al., 2022 (IEEE/CVF CVPR) [39]<br><br>Source code and model: https://huggingface.co/pamixsun/swinv2_tiny_for_glaucoma_classification |
| SegFormer-B0 | SegFormer (Lightweight hierarchical encoder with MLP decoder) | Multi-class semantic segmentation of the optic disc and cup | REFUGE Challenge dataset (n = 1,200$) | Color fundus photograph (RGB) | Pixel-wise segmentation mask (3 classes); vertical cup-to-disc ratio (vCDR) | base architecture via Xie et al., 2021 (NeurIPS) [40]<br><br>Source code and model: https://huggingface.co/pamixsun/segformer_for_optic_disc_cup_segmentation |

Table 2. Glaucoma detection performance versus dataset labels across LLM backbones, image preprocessing, and workflow configurations (Run 1)

| Dataset | Preprocessing | Configuration | Gemini | | | | ChatGPT | | | |
|---|---|---|---|---|---|---|---|---|---|---|
| | | | Accuracy | Sensitivity | Specificity | F1 | Accuracy | Sensitivity | Specificity | F1 |
| **ORIGA** | Uncropped | LLM alone | 0.57 | 0.40 | 0.74 | 0.48 | 0.62 | 0.84 | 0.40 | 0.69 |
| | | Agentic AI | 0.73 | 0.76 | 0.70 | 0.74 | 0.72 | 0.90 | 0.54 | 0.76 |
| | Cropped | LLM alone | 0.55 | 0.72 | 0.38 | 0.62 | 0.50 | 1.00 | 0.00 | 0.67 |
| | | Agentic AI | 0.73 | 0.74 | 0.72 | 0.73 | 0.69 | 0.98 | 0.40 | 0.76 |
| | Reference * standard | *Specialist* | *0.79* | *0.76* | *0.83* | *0.79* | - | - | - | - |
| **RIM-ONE v3** | Uncropped | LLM alone | 0.61 | 0.87 | 0.44 | 0.64 | 0.38 | 0.97 | 0.00 | 0.55 |
| | | Agentic AI | 0.85 | 0.80 | 0.89 | 0.81 | 0.85 | 0.77 | 0.90 | 0.80 |
| | Cropped | LLM alone | 0.54 | 0.82 | 0.36 | 0.58 | 0.40 | 0.95 | 0.05 | 0.55 |
| | | Agentic AI | 0.88 | 0.79 | 0.93 | 0.83 | 0.87 | 0.79 | 0.92 | 0.83 |
| | Reference * standard | *Specialist* | *0.88* | *0.85* | *0.90* | *0.85* | - | - | - | - |

***Reference standard:** The fellowship-trained glaucoma specialist achieved accuracy 0.79, sensitivity 0.76, specificity 0.83, F1 0.79 on ORIGA and accuracy 0.88, sensitivity 0.85, specificity 0.90, F1 0.85 on RIM-ONE-v3. Specialist values are dataset-level reference standards independent of model, preprocessing, or workflow condition

Table 3. Agreement with the glaucoma specialist across two repeated runs and run-to-run consistency, across LLM backbones, image preprocessing, and workflow configurations.

| Dataset | Preprocessing | Configuration | Gemini | | | ChatGPT | | |
|---|---|---|---|---|---|---|---|---|
| | | | Run 1 vs. specialist | Run 2 vs. specialist | Run 1 vs. Run 2 | Run 1 vs. specialist | Run 2 vs. specialist | Run 1 vs. Run 2 |
| **ORIGA** | Uncropped | LLM alone | 0.30 | 0.36 | 0.34 | 0.20 | 0.21 | 0.51 |
| | | Agentic AI | 0.51 | 0.55 | 0.54 | 0.48 | 0.40 | 0.72 |
| | Cropped | LLM alone | 0.07 | 0.21 | 0.44 | 0.00 | 0.00 | 0.00 |
| | | Agentic AI | 0.38 | 0.51 | 0.40 | 0.39 | 0.40 | 0.91 |
| **RIM-ONE v3** | Uncropped | LLM alone | 0.25 | 0.17 | 0.42 | -0.02 | 0.01 | -0.01 |
| | | Agentic AI | 0.81 | 0.80 | 0.87 | 0.78 | 0.77 | 0.85 |
| | Cropped | LLM alone | 0.22 | 0.11 | 0.47 | -0.03 | 0.01 | 0.42 |
| | | Agentic AI | 0.84 | 0.80 | 0.93 | 0.83 | 0.78 | 0.96 |

**All values are Cohen's κ.**
"Run 1 vs. specialist" and "Run 2 vs. specialist" = agreement of each independent API run with the glaucoma specialist; "Run 1 vs. Run 2" = run-to-run agreement between the two repeated runs of the same predictor on the same images.
Per Landis–Koch: <0.2 slight, 0.2–0.4 fair, 0.4–0.6 moderate, 0.6–0.8 substantial, >0.8 almost perfect.

Table 4. Cup-to-disc ratio estimation versus the glaucoma specialist, across LLM backbones, image preprocessing, and workflow configurations.

| Dataset | Preprocessing | Configuration | Gemini | | | | ChatGPT | | | |
|---|---|---|---|---|---|---|---|---|---|---|
| | | | Categorical | | Numerical | | Categorical | | Numerical | |
| | | | Acc | F1 | MAE | Pearson r | Acc | F1 | MAE | Pearson r |
| ORIGA | Uncropped | LLM alone | 0.64 | 0.63 | 0.178 | 0.35 | 0.65 | 0.75 | 0.156 | 0.28 |
| | | Agentic AI | 0.77 | 0.79 | 0.132 | 0.69 | 0.78 | 0.83 | 0.132 | 0.62 |
| | Cropped | LLM alone | 0.62 | 0.71 | 0.166 | 0.21 | 0.61 | 0.76 | 0.159 | 0.12 |
| | | Agentic AI | 0.73 | 0.73 | 0.122 | 0.70 | 0.78 | 0.85 | 0.126 | 0.59 |
| RIM-ONE v3 | Uncropped | LLM alone | 0.57 | 0.64 | 0.187 | 0.39 | 0.46 | 0.63 | 0.228 | 0.35 |
| | | Agentic AI | 0.90 | 0.88 | 0.104 | 0.84 | 0.78 | 0.80 | 0.114 | 0.82 |
| | Cropped | LLM alone | 0.56 | 0.62 | 0.205 | 0.28 | 0.46 | 0.63 | 0.223 | 0.14 |
| | | Agentic AI | 0.87 | 0.83 | 0.114 | 0.81 | 0.76 | 0.77 | 0.114 | 0.83 |

Categorical CDR = enlarged vs. normal compared with the glaucoma specialist's categorical grade. Numerical CDR: MAE = mean absolute error in CDR units (range 0–1); Pearson r = correlation between predicted and specialist numerical CDR.

Table 5. Three-class image quality assessment (good / moderate / poor) versus the glaucoma specialist and run-to-run consistency, across LLM backbones, image preprocessing, and workflow configurations.

| Dataset | Preprocessing | Configuration | Gemini | | | | ChatGPT | | | |
|---|---|---|---|---|---|---|---|---|---|---|
| | | | vs. specialist | | | Run-to-run | vs. specialist | | | Run-to-run |
| | | | Acc | Quad. κ | Poor recall | Quad. κ | Acc | Quad. κ | Poor recall | Quad. κ |
| ORIGA | Uncropped | LLM alone | 0.43 | 0.11 | 0% (0/13) | 0.53 | 0.52 | 0.40 | 8% (1/13) | 0.87 |
| | | Agentic AI | 0.50 | 0.41 | 31% (4/13) | 0.85 | 0.51 | 0.40 | 23% (3/13) | 0.81 |
| | Cropped | LLM alone | 0.52 | 0.47 | 54% (7/13) | 0.92 | 0.49 | 0.30 | 69% (9/13) | 0.65 |
| | | Agentic AI | 0.54 | 0.50 | 62% (8/13) | 0.92 | 0.52 | 0.32 | 77% (10/13) | 0.76 |
| RIM-ONE v3 | Uncropped | LLM alone | 0.50 | 0.42 | 56% (10/18) | 0.76 | 0.70 | 0.45 | 94% (17/18) | 0.73 |
| | | Agentic AI | 0.39 | 0.10 | 100% (18/18) | 0.53 | 0.41 | 0.10 | 100% (18/18) | 0.60 |
| | Cropped | LLM alone | 0.38 | 0.37 | 82% (14/17) | 0.83 | 0.38 | 0.10 | 94% (17/18) | 0.52 |
| | | Agentic AI | 0.30 | 0.09 | 100% (17/17) | 0.56 | 0.29 | 0.08 | 100% (18/18) | 0.31 |

Quad. κ = quadratic-weighted Cohen's κ, which accounts for the ordered structure of the three quality categories by penalizing disagreements according to their distance. Poor recall = proportion of specialist-graded "poor" images correctly identified as poor by the predictor. Run-to-run Quad. κ = agreement between the two independent API runs of the same predictor on the same images. N = 100 per dataset; specialist quality distributions are 35 good / 52 moderate / 13 poor (ORIGA) and 11 good / 71 moderate / 18 poor (RIM-ONE v3).

For RIM-ONE-v3 Cropped conditions, the denominator for poor-quality recall is 17 rather than 18 in select configurations, reflecting the exclusion of one image that was classified as ungradable by the predictor and therefore not assigned a quality grade of good, moderate, or poor.

Supplemental Table S1. Cup-to-disc ratio run-to-run consistency between two repeated runs, across LLM backbones, image preprocessing, and workflow configurations.

| Dataset | Preprocessing | Configuration | Gemini | | ChatGPT | |
|---|---|---|---|---|---|---|
| | | | Categorical κ | Numerical MAE | Categorical κ | Numerical MAE |
| ORIGA | Uncropped | LLM alone | 0.33 | 0.141 | 0.66 | 0.053 |
| | | Agentic AI | 0.58 | 0.053 | 0.68 | 0.049 |
| | Cropped | LLM alone | 0.42 | 0.110 | 0.00 | 0.012 |
| | | Agentic AI | 0.51 | 0.047 | 0.73 | 0.051 |
| RIM-ONE v3 | Uncropped | LLM alone | 0.34 | 0.103 | 0.00 | 0.019 |
| | | Agentic AI | 0.89 | 0.037 | 0.59 | 0.027 |
| | Cropped | LLM alone | 0.47 | 0.125 | 0.49 | 0.021 |
| | | Agentic AI | 0.90 | 0.019 | 0.65 | 0.010 |

Categorical κ = Cohen's κ between the two runs' enlarged/normal classifications; Numerical MAE = mean absolute error between the two runs' numerical CDR values. Run-to-run consistency reflects the agreement between two independent API runs of the same predictor on the same images.

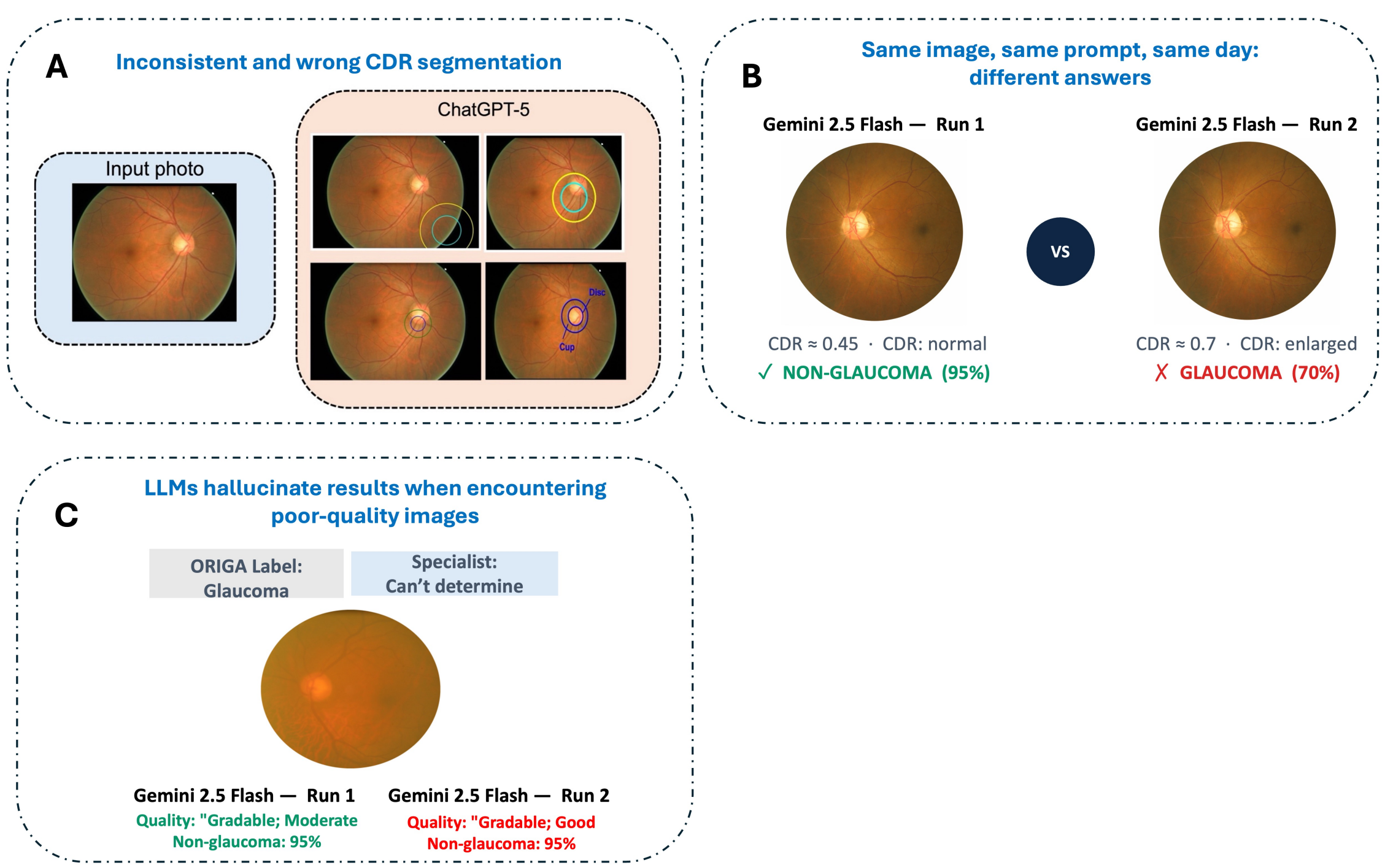

**Figure 1:** Fundamental limitations of large language models for glaucoma assessment from fundus photographs. (A) Inconsistent and inaccurate cup-to-disc ratio (CDR) segmentation by ChatGPT-5: the same fundus photograph analyzed across four repeated runs produces markedly different optic disc and cup boundary delineations. (B) Stochastic diagnostic variability in Gemini 2.5 Flash: identical image and prompt submitted on the same day yield contradictory outputs across two runs. (C) Hallucination in the presence of a poor-quality fundus image: despite the image being ungradable by a specialist, the LLM generates confident diagnostic outputs across both runs, assigning a gradable quality rating and a non-glaucoma classification with 95% confidence, rather than appropriately flagging the image as insufficient for interpretation.

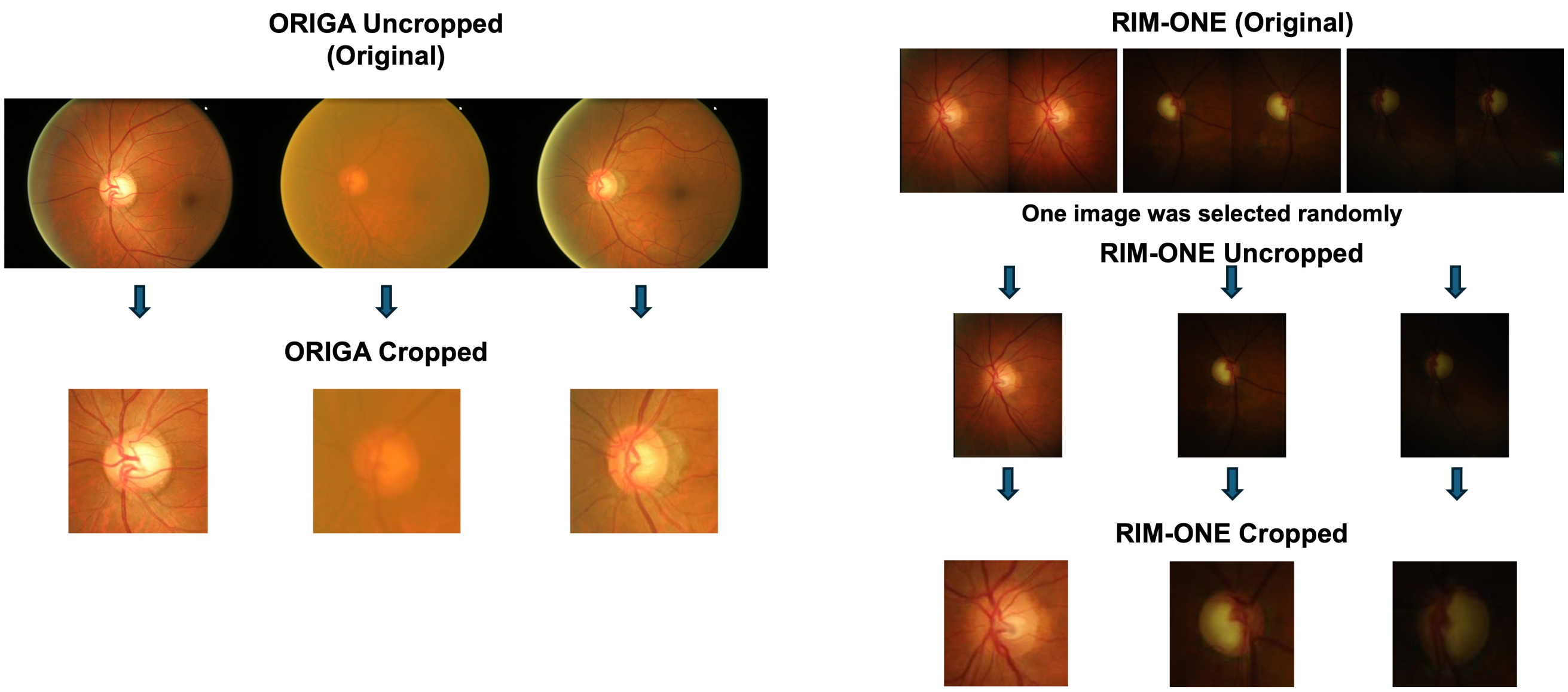

**Figure 2:** Representative fundus photographs before and after optic disc-centered cropping from the ORIGA and RIM-ONE datasets (uncropped and cropped versions), illustrating variability in illumination and image quality across examples.

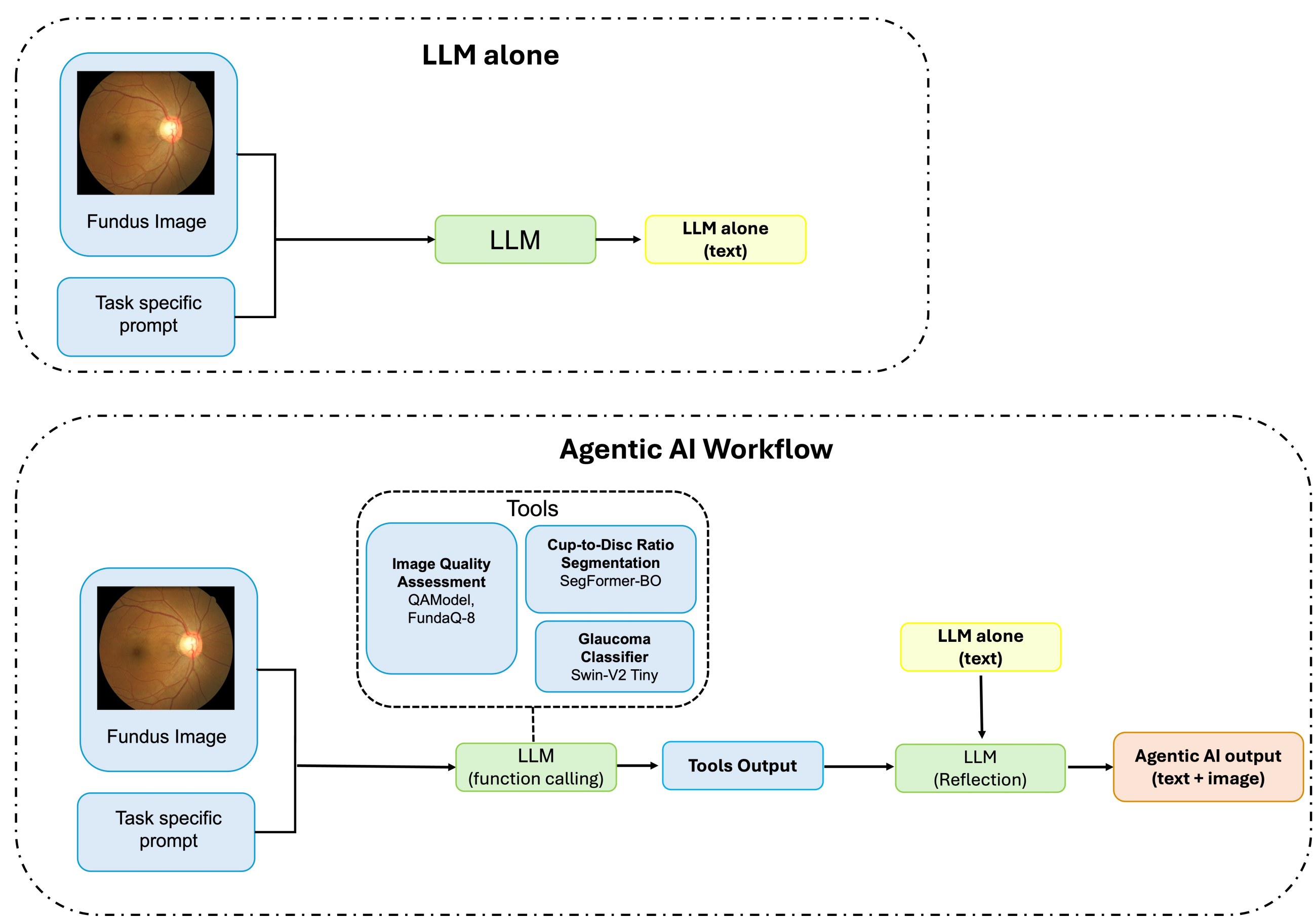

**Figure 3:** Comparison of the LLM-alone and agentic AI workflows for fundus-based glaucoma assessment. In the LLM-alone workflow (top), a fundus image and task-specific prompt are passed directly to the LLM, producing a text-only diagnostic output. In the agentic workflow (bottom), the LLM first uses function calling to invoke four specialized deep learning tools, image quality assessment (QAModel, FundaQ-8), cup-to-disc ratio segmentation (SegFormer-B0), and glaucoma classification (SwinV2-Tiny), and then applies reflection to integrate the tool outputs with its initial impression, producing a final diagnostic report combining text and an annotated image.

**A**

Agentic AI refines the incorrect initial LLM prediction by reflecting on the outputs of the DL models.

Input

ORIGA Label: Glaucoma

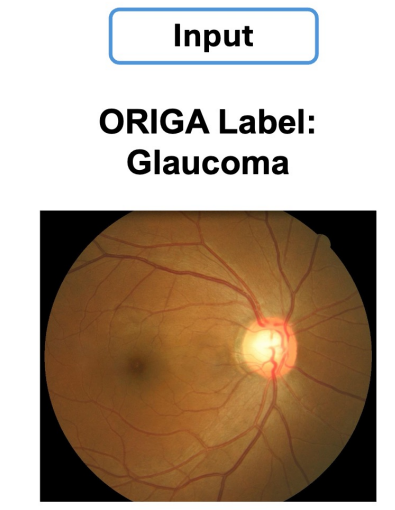

LLM Alone

1. Image Quality Assessment:
   • Gradability: Gradable
   • Quality level: Good
2. Optic Disc / Cup Assessment:
   • Cup-to-Disc Ratio (CDR): Normal
   • Cup-to-Disc Ratio (CDR): 0.45
3. Glaucoma Status:
   • Patient status: Non-Glaucoma
   • Provide a probability score for each class:
   - P(Glaucoma) = 0.10 - P(Non-Glaucoma) = 0.90

Agentic AI

1. Image Quality Assessment:
   • Gradability: Gradable
   • Quality level: Good
2. Optic Disc / Cup Assessment:
   • Cup-to-Disc Ratio (CDR): Normal
   • Cup-to-Disc Ratio (CDR): 0.61
3. Glaucoma Status:
   • Patient status: Glaucoma
   • Provide a probability score for each class:
   - P(Glaucoma) = 0.98 - P(Non-Glaucoma) = 0.02

**Reasoning:** Both analyses agree on the image being of good quality and gradable. However, there is a substantial disagreement regarding the Cup-to-Disc Ratio (CDR) and the overall glaucoma status. The initial LLM estimated a normal CDR of 0.45, leading to a "Non-Glaucoma" diagnosis with high confidence. In contrast, the specialized deep learning pipeline, including a dedicated segmentation model (SegFormer-B0), estimated an enlarged CDR of 0.61, and its glaucoma classifier (SwinV2-Tiny) predicted "Glaucoma" with very high confidence (99.5%). As a glaucoma specialist, I find the higher CDR estimate from the segmentation model, which is consistent with the deep learning classifier's strong glaucoma prediction, to be more clinically significant and reliable. Therefore, I prioritize the findings of the specialized deep learning models, particularly the enlarged CDR, which points towards a higher likelihood of glaucoma.

**B**

Agentic AI confirms the correct initial LLM prediction by reflecting on the outputs of the DL models.

Input

ORIGA Label: Non-Glaucoma

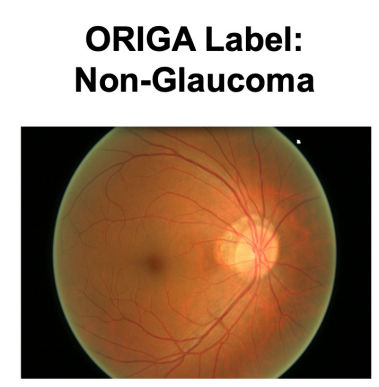

LLM Alone

1. Image Quality Assessment:
   • Gradability: Gradable
   • Quality level: Good
2. Optic Disc / Cup Assessment:
   • Cup-to-Disc Ratio (CDR): Normal
   • Cup-to-Disc Ratio (CDR): 0.5
3. Glaucoma Status:
   • Patient status: Non-Glaucoma
   • Provide a probability score for each class:
   - P(Glaucoma) = 0.10 - P(Non-Glaucoma) = 0.90

Agentic AI

1. Image Quality Assessment:
   • Gradability: Gradable
   • Quality level: Good
2. Optic Disc / Cup Assessment:
   • Cup-to-Disc Ratio (CDR): Normal
   • Cup-to-Disc Ratio (CDR): 0.4
3. Glaucoma Status:
   • Patient status: Non-Glaucoma
   • Provide a probability score for each class:
   - P(Glaucoma) = 0.2 - P(Non-Glaucoma) = 0.8

**Reasoning:** Integrated my re-evaluation with both analyses. All sources concurred on good image quality and gradability. My own assessment and the Deep Learning SegFormer-B0 model's estimated vertical Cup-to-Disc Ratio (vCDR) of 0.38 aligned closely, indicating a small cup, which was also reflected in the initial LLM's 0.5 CDR estimate. Morphologically, the optic disc shows a healthy, well-perfused neuroretinal rim with preserved ISNT rule and no signs of glaucomatous damage. Despite the SwinV2-Tiny DL model predicting high glaucoma probability (0.88), this finding strongly contradicts the clear morphological evidence from direct observation, the small CDR from the segmentation model, and the initial LLM's assessment. Therefore, I prioritize the robust morphological findings and the agreement across multiple data points (my assessment, LLM, and DL CDR) that indicate a non-glaucomatous disc.

**Figure 4:** Agentic workflow error correction demonstrating bidirectional correction. (A) Correction of LLM error using DL tool outputs. Ground truth: glaucoma. Gemini alone estimated a normal CDR (0.40) and classified the image as non-glaucoma. The segmentation model (SegFormer-B0) identified an enlarged CDR (0.86) and the classifier (SwinV2-Tiny) returned 99.5% glaucoma probability. During reflection, the LLM integrated this evidence and correctly reclassified the image as glaucoma. (B) Resolution of conflicting DL outputs through clinical reasoning. Ground truth: non-glaucoma. The classifier predicted high glaucoma probability (88%), but the segmentation model estimated a normal CDR (0.38). During reflection, the LLM recognized this discrepancy and prioritized the morphological evidence, the normal cup-to-disc ratio consistent with its initial assessment, correctly maintaining the non-glaucoma classification. These examples demonstrate that the agentic workflow functions as an intelligent arbitrator, leveraging tool agreement to correct LLM errors while applying clinical reasoning to resolve tool disagreement.

**LLM-alone prompt**

```
You are a glaucoma specialist. Analyze this fundus image
and provide:

1. Image Quality Assessment:
   - Gradability: "Gradable" or "Ungradable"
   - Quality level: "Good", "Moderate", or "Poor"

2. Optic Disc / Cup Assessment:
   - Cup-to-Disc Ratio (CDR): "Normal" or "Enlarged"
   - Cup-to-Disc Ratio (CDR): please provide a numerical
estimate (e.g., 0.5 or 0.6)

3. Glaucoma Status:
   - Patient status: "Non-Glaucoma" or "Glaucoma"
   - Provide a probability score for each class:
       - P(Glaucoma) = X
       - P(Non-Glaucoma) = Y

Don't explain! be straight forward just answering these
questions!
```

**Agentic AI prompt**

```
You are a glaucoma specialist integrating two independent analyses
of the SAME fundus image (which I'm showing you now).

=== Analysis #1: One LLM Initial Response ===
{initial_text}

=== Analysis #2: Deep Learning Pipeline Report from some pretrained models ===
{dl_text}

============================
TASK:
You must re-evaluate the fundus image yourself during this reflection step.
Then, using your new image assessment together with the two previous analyses,
produce ONE unified final clinical conclusion.

You must integrate three sources of information:
 1. Your own fresh re-evaluation of the image (look at the image I'm showing you)
 2. The initial LLM output
 3. The deep learning (DL) model outputs

Your goal is to reconcile all three perspectives and produce your single
best expert judgment on this image.

Your final output MUST contain exactly TWO sections:

1. Final Answers to the Original Prompt:
   Provide a structured, direct answer that includes:
     - Gradability (Gradable / Ungradable)
     - Image quality (Good / Moderate / Poor)
     - Cup-to-Disc Ratio category (Normal / Enlarged)
     - Numerical CDR estimate
     - Glaucoma status (Non-Glaucoma / Glaucoma)
     - Probability scores for both classes:
         - P(Glaucoma) = X
         - P(Non-Glaucoma) = Y

2. Reasoning Paragraph (Short):
   Provide one concise/short paragraph explaining how you integrated the Gemini interpretation and
the DL model findings, and how this led to your final decision. just mention important things.

IMPORTANT:
Your final answers in Section 2 should represent your highest-confidence interpretation after
integrating all available information.
```

**Supplemental Figures S1:** Supplemental Figure S1. Prompt templates used for the LLM-alone and agentic AI workflows. (Left) The LLM-alone prompt requesting structured assessment of image quality, optic disc/cup features, and glaucoma status directly from the fundus image. (Right) The agentic AI reflection prompt, in which the LLM integrates its own fresh re-evaluation of the image with the initial LLM analysis and deep learning pipeline outputs to produce a unified final clinical conclusion.

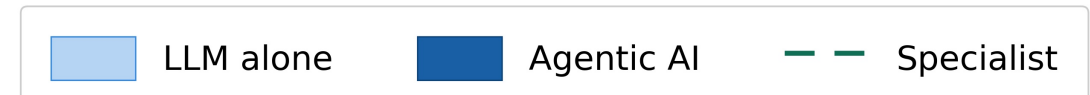

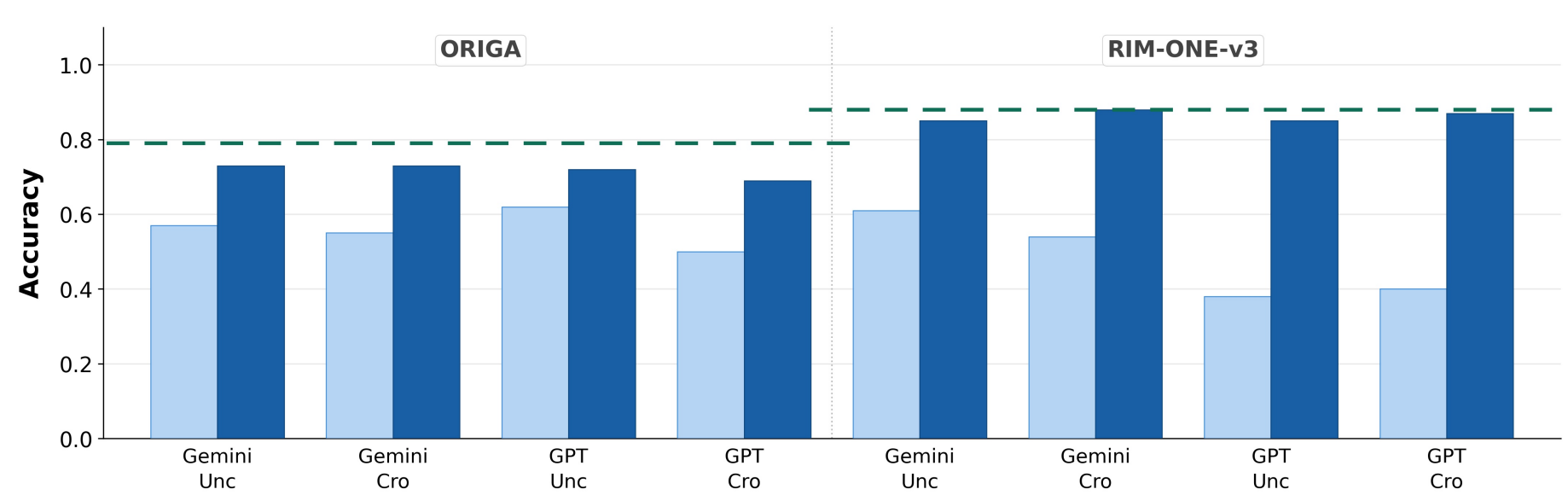

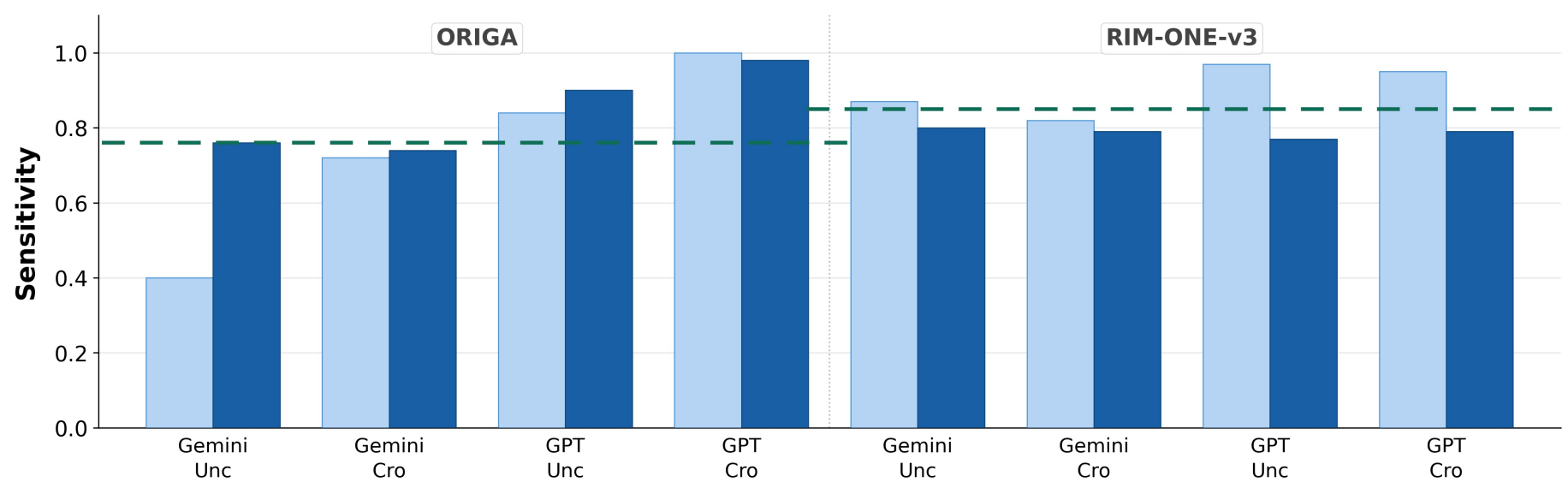

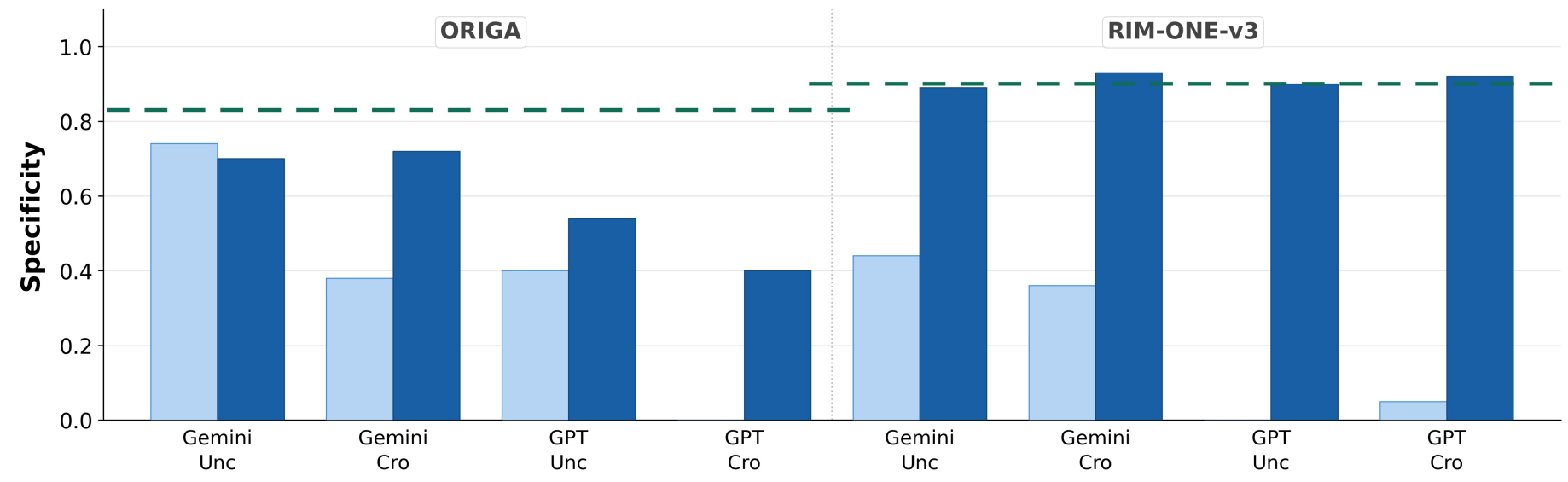

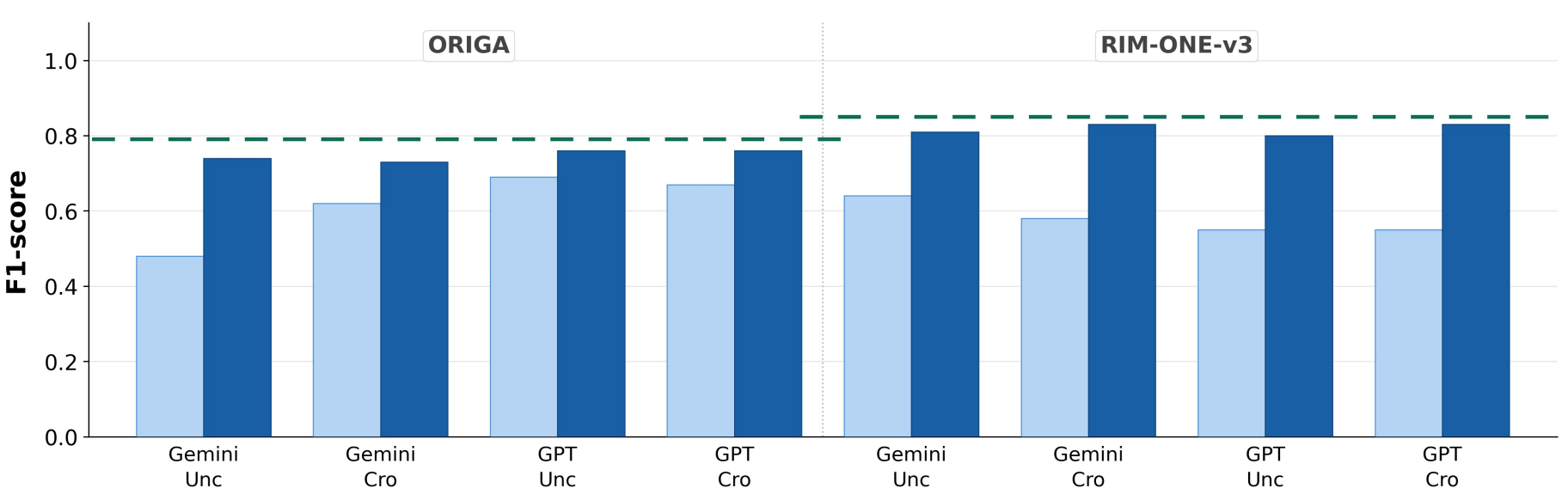

**Supplemental Figures S2:** Glaucoma classification performance of LLM-alone and agentic AI workflows across all experimental conditions. Accuracy, sensitivity, specificity, and F1-score for Gemini 2.5 Flash and GPT-5.4 mini under LLM-alone and agentic AI workflows, evaluated on the ORIGA and RIM-ONE-v3 datasets with uncropped (Unc) and cropped (Cro) fundus images (Run 1). Dashed lines indicate the fellowship-trained glaucoma specialist's performance for each dataset. The agentic workflow consistently improved all four metrics across both LLMs and both datasets, with the most pronounced gains in specificity for GPT-5.4 mini, which exhibited near-zero specificity under the LLM-alone workflow across three of four conditions

Input

Output of DL tools

ORIGA Label: Glaucoma

SegFormer Overlay (vCDR=0.61)

**Unified Glaucoma Analysis Report:**

Image Quality (2-model evaluation):
• QAModel (Inception-V3 quality CNN): Poor (score 0.32)
• FundaQ-8 (8-parameter clinical quality model): Poor (score 0.43)

Deep Learning Models
• SwinV2-Tiny — glaucoma probability classifier trained on REFUGE: → Prediction: Glaucoma (confidence 0.89) → P(glaucoma) = 88.7% | P(Non-Glaucoma) = 11.3%

• SegFormer-B0 — optic disc/cup segmentation model: → Estimated vCDR = 0.61

ORIGA Label: Non-Glaucoma

SegFormer Overlay (vCDR=0.44)

**Unified Glaucoma Analysis Report:**
Image Quality (2-model evaluation)
• QAModel (Inception-V3 quality CNN): Good (score 1.00)
• FundaQ-8 (8-parameter clinical quality model): Moderate (score 0.84)

Deep Learning Models
• SwinV2-Tiny — glaucoma probability classifier trained on REFUGE: → Prediction: Non-Glaucoma (confidence 0.97) → P(glaucoma) = 3.4% | P(Non-Glaucoma) = 99.6%

• SegFormer-B0 — optic disc/cup segmentation model: → Estimated vCDR = 0.44

RIM-ONE v3 Label: Non-Glaucoma

SegFormer Overlay (vCDR=0.43)

**Unified Glaucoma Analysis Report:**

Image Quality (2-model evaluation):
• QAModel (Inception-V3 quality CNN): Poor (score 0.40)
• FundaQ-8 (8-parameter clinical quality model): Poor (score 0.44)

Deep Learning Models
• SwinV2-Tiny — glaucoma probability classifier trained on REFUGE: → Prediction: Non-Glaucoma (confidence 1.00) → P(glaucoma) = 0.4% | P(Non-Glaucoma) = 99.6%

• SegFormer-B0 — optic disc/cup segmentation model: → Estimated vCDR = 0.43

RIM-ONE v3 Label: Glaucoma

SegFormer Overlay (vCDR=0.71)

**Unified Glaucoma Analysis Report:**
Image Quality (2-model evaluation)
• QAModel (Inception-V3 quality CNN): Poor (score 0.00)
• FundaQ-8 (8-parameter clinical quality model): Poor (score 0.00)

Deep Learning Models
• SwinV2-Tiny — glaucoma probability classifier trained on REFUGE: → Prediction: Glaucoma (confidence 0.97) → P(glaucoma) = 97.5% | P(Non-Glaucoma) = 2.5%

• SegFormer-B0 — optic disc/cup segmentation model: → Estimated vCDR = 0.71

**Supplemental Figures S3:** Representative outputs of the four deep learning tools for one example image from each dataset/glaucoma-status combination (ORIGA glaucoma, ORIGA non-glaucoma, RIM-ONE-v3 non-glaucoma, RIM-ONE-v3 glaucoma), including quality scores, glaucoma probability, and optic disc/cup segmentation with estimated CDR.

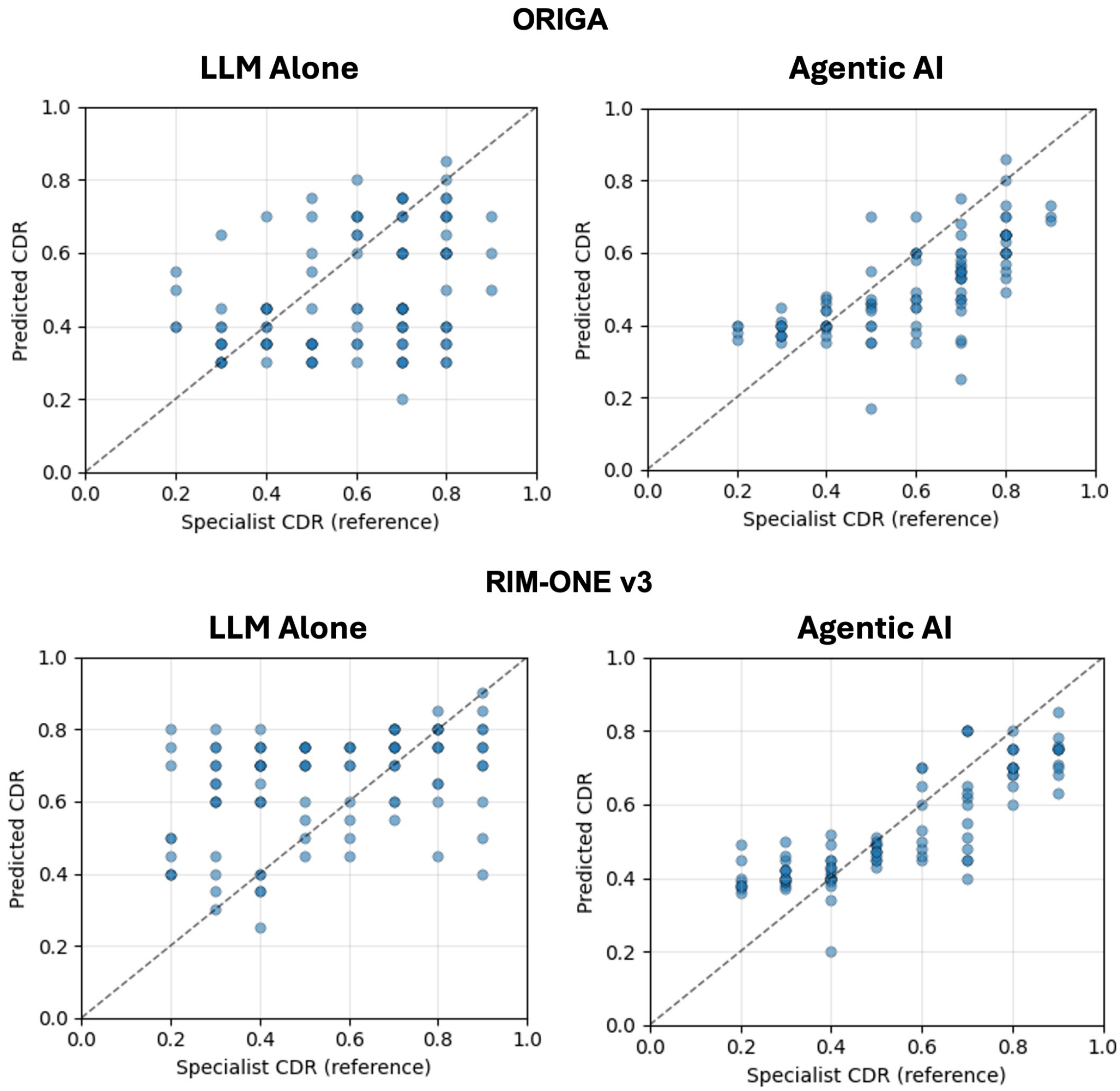

**Supplemental Figures S2:** Predicted versus specialist CDR for Gemini 2.5 Flash on uncropped ORIGA and RIM-ONE-v3 images, comparing LLM-alone and agentic AI workflows. The dashed line represents perfect agreement; agentic predictions cluster more closely around it than LLM-alone predictions.